\documentclass[10pt]{article} 
\usepackage[preprint]{rlc}

\usepackage{amssymb}            
\usepackage{mathtools}          
\usepackage{mathrsfs}           
\mathtoolsset{showonlyrefs}     
\usepackage{graphicx}           
\usepackage{subcaption}         
\usepackage[space]{grffile}     
\usepackage{url}                

\usepackage{amsmath}  

\usepackage{algorithm}  
\usepackage{algorithmic}
\usepackage{booktabs}    
\usepackage{multirow}    
\usepackage{tabularx}    
\usepackage{makecell}    
\usepackage{amssymb}     
\usepackage{enumerate}
\usepackage{enumitem} 
\usepackage{wrapfig}

\usepackage[most]{tcolorbox}
\tcbuselibrary{skins, breakable}
\usepackage{xcolor}
\tcbset{
  promptstyle/.style={
    breakable,
    enhanced,
    colback=gray!4,
    colframe=gray!60!black,
    boxrule=0.6pt,
    arc=2pt,
    left=8pt, right=8pt, top=18pt, bottom=6pt,
    fontupper=\small\ttfamily,
    before upper={\parindent0pt \linespread{0.95}\selectfont},
    attach boxed title to top left={yshift=-5.6mm, xshift=0.4mm},
    coltitle=black,
    fonttitle=\bfseries\sffamily\footnotesize,
    boxed title style={
      colback=gray!20,
      colframe=gray!40,
      arc=1pt,
      outer arc=0pt,
      boxrule=0.4pt
    }
  }
}

\newtcolorbox[auto counter, number within=section, number format=\arabic]{promptbox}[2][]{
  promptstyle,
  title={Prompt~\thetcbcounter: #2},
  #1
}

\title{Retrieval-Driven Memory Reconsolidation for Long-Term LLM Agents}

\author{\\
\name{Yuanyi Song}$^{1}$\thanks{This work was done during Yuanyi Song’s internship at OPPO.}, 
\name{Yukai Wang}$^{1}$,
\name{Xinbei Ma}$^{1,3}$,
\name{Zhihui Fu}$^{2}$,
\name{Jianghao Lin}$^{1}$\textsuperscript{†},
\name{Weiwen Liu}$^{1}$\textsuperscript{†},\\
\name{Jun Wang}$^{2}$\textsuperscript{†},
\name{Huarong Deng}$^{2}$,
\name{Yong Yu}$^{1}$,
\name{Weinan Zhang}$^{1}$\thanks{J. Lin, W. Liu, J. Wang and W. Zhang are the corresponding authors.}\\
$^1$Shanghai Jiao Tong University \quad
$^2$OPPO
\quad $^3$National University of Singapore\\
\texttt{norsheep919@sjtu.edu.cn}\quad \texttt{linjianghao@sjtu.edu.cn}\\
\texttt{wwliu@sjtu.edu.cn}\quad
\texttt{wnzhang@sjtu.edu.cn}\\
}

\begin{document}

\maketitle

\begin{abstract}
Long-term memory is essential for LLM-based agents operating over extended interactions. 
Existing memory systems primarily update memory when new information arrives, treating retrieval as the endpoint of memory access rather than a driver of memory evolution. 
Consequently, retrieval feedback is rarely exploited to reorganize memory for future access continuously. Moreover, most existing approaches rely on predefined memory structures together with fixed retrieval pipelines, limiting the agent's ability to organize and evolve its own memory autonomously.
Inspired by memory reconsolidation in cognitive neuroscience, we propose \textbf{REALM}, a \textbf{r}econsolidation-\textbf{e}volution \textbf{a}gentic \textbf{l}ong-term \textbf{m}emory framework. It models long-term memory as a continual lifecycle by autonomously organizing memories into a heterogeneous cognitive graph, retrieving evidence via adaptively composed graph-search atoms, and continually reconsolidating memories based on retrieval feedback.
REALM achieves an average accuracy of 75.97\% on LoCoMo and 65.11\% on LongMemEval, outperforming the strongest baselines by 7.17 and 1.31 points respectively.
Ablation studies confirm that memory reconsolidation consistently boosts performance, with further analyses revealing that it progressively reorganizes related memory units into more coherent local structures for collective evidence recall and utilization during reasoning.
These results suggest that retrieval-driven memory reconsolidation provides an effective mechanism for continually evolving long-term memory in LLM agents.
\end{abstract}

\section{Introduction}
\label{sec:introduction}

Recent advances in Large Language Model (LLM)-based agents have enabled their deployment in long-term interactive scenarios like personal assistants, long-horizon task collaboration, and continual decision-making~\citep{zhang2025survey, du2026memory,song2026colorbench}. 
In these scenarios, agents must accumulate knowledge across extensive histories and leverage past experiences for future reasoning, making long-term memory a foundational capability of intelligent agents~\citep{wang2024memoryllm,li2026cam,fang2026memp,chai2026smmbench}.  

\begin{figure}[t]
    \centering
    \includegraphics[width=0.8\linewidth]{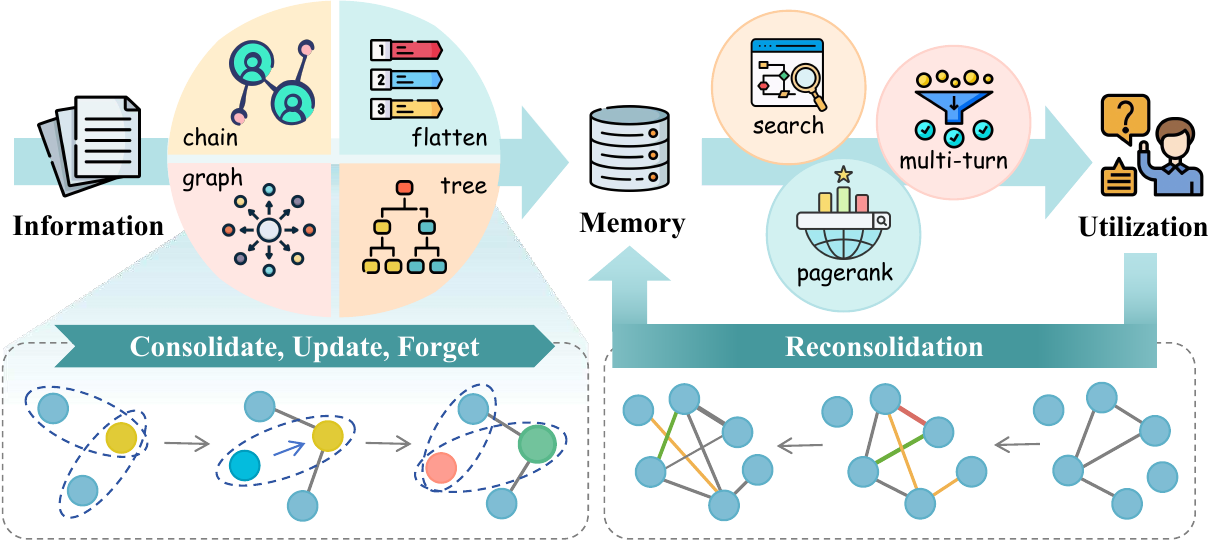}
    \caption{\textbf{Motivation of retrieval-driven memory reconsolidation.} Existing memory systems mainly update memories with new information using predefined data structures and a fixed retrieval pipeline, while REALM forms a closed-loop memory lifecycle, leveraging retrieval feedback to dynamically reorganize the memory topology upon utilization.}
    \label{fig:introduction}
\vspace{-1em}
\end{figure}

As shown in the left part of Figure~\ref{fig:introduction}, existing long-term memory research predominantly follows a forward evolution paradigm, where memory updates, including addition, deletion, and merging of memory units, are triggered exclusively by newly acquired information~\citep{zhong2024memorybank,hu2025memory,wang2025mirix}.  
Under this paradigm, memory retrieval is treated as a passive, terminal endpoint of utilization rather than a catalyst for structural refinement.

This prevailing paradigm stands in sharp contrast to memory reconsolidation in cognitive neuroscience, a foundational mechanism where recalled memory is not static data~\citep{adam2026eval,lee2025memory,samieiyeganeh2026forgetting}; instead, reactivation initiates a labile state where connections are strengthened, weakened, or newly established, allowing memory networks to actively evolve through usage and enabling them to remain adaptable to new experiences while maintaining stability~\citep{nader2000fear}. 
Driven by this, long-term memory should be a dynamic memory lifecycle that possesses the feedback-driven autonomy to realign its organization after each retrieval. However, achieving such post-retrieval evolution requires balancing a highly flexible memory structure with an adaptive retrieval mechanism over evolving topologies~\citep{hu2026does}.

To address this challenge, we propose \textbf{REALM}, \textbf{a \underline{r}econsolidation-\underline{e}volution \underline{a}gentic \underline{l}ong-term \underline{m}emory lifecycle framework}, which instantiates this holistic lifecycle through three complementary components:
(1) Autonomous Organization: We define atomic memory nodes and relations, empowering agents to autonomously construct local topologies that organically emerge into a unified agentic cognitive graph. 
(2) Adaptive Retrieval: We decompose complex graph traversal into atomic search actions, allowing agents to compose context-tailored retrieval policies dynamically.
(3) Memory Reconsolidation: After each task, we perform local reconsolidation on the activated subgraph, dynamically updating connections to achieve continually retrieval-driven evolution.
Experiments on standard benchmarks demonstrate that REALM consistently outperforms conventional post-static paradigms, with further analyses confirming that memory reconsolidation effectively optimizes memory structures over long horizons. 
At the same time, the agentic cognitive graph enables efficient memory organization and retrieval. 
These findings support our hypothesis that long-term memory should continuously evolve through usage rather than merely accumulate new information.

Our main contributions are summarized as follows:
\begin{itemize}
    \item \textbf{Memory Lifecycle Concept.} We introduce a lifecycle perspective for long-term memory in LLM agents, where memory continuously evolves through usage rather than remaining static after construction.
    \item \textbf{REALM Framework.} We propose a unified agentic memory framework that integrates autonomous memory organization, adaptive retrieval, and post-retrieval reconsolidation into a unified evolution lifecycle.
    \item \textbf{Empirical Validation.} Comprehensive evaluations show that lifecycle-driven memory evolution consistently improves long-term memory performance, revealing the autonomous emergence of reusable memory topologies and retrieval policies.
\end{itemize}

\section{Related Work}
\label{sec:related-work}

\subsection{Agent Long-term Memory}
Existing long-term memory systems mainly focus on memory organization and retrieval, continuously accumulating, organizing, and utilizing historical information~\citep{langmem2025,zhou2026externalization}.
Early works rely on flat incremental stores to extract and summarize historical interactions~\citep{zhong2024memorybank,chhikara2025mem0,wang2025m+}. To enhance representation, recent approaches adopt explicit structures, including hierarchical trees or mind maps~\citep{rezazadeh2025isolatedmemtree,li2026cam,xu2026amem}, knowledge graphs~\citep{edge2024localgraphrag,yang2026plugmem}, temporal~\citep{rasmussen2025zep} or OS-inspired management~\citep{li2025memos,packer2023memgpt,kang2025memoryos,zhu2026lossy}, and multimodal representations~\citep{wang2025mirix}. Correspondingly, retrieval research enhances utilization via multi-hop graph reasoning~\citep{gutierrez2024hipporag}, iterative search~\citep{yan2025gam,du2025memr}, dual-process retrieval~\citep{you2026d} and adaptive graph traversal~\cite{jiang2026magma}. However, these methods typically adhere to predefined organizations with static post-retrieval states. In contrast, we formulate a unified memory lifecycle that augments organization and retrieval with dynamic, autonomous memory reorganization.

\subsection{Agent Memory Evolution and Lifecycle} 
Inspired by cognitive neuroscience, incorporating memory stages like encoding and consolidation has gained traction~\citep{gutierrez2024hipporag,khiste2014mem,jiang2026synapse}. Neuroscience reveals that recalled memories become transiently labile and undergo reconsolidation for continual adaptation~\citep{nader2000labile}, implying that long-term memory evolves through repeated use rather than one-time storage~\citep{gonzalez2026subspace}. While several studies explore memory evolution via offline compression~\citep{fang2025lightmem}, recency and frequency heuristics~\citep{zhong2024memorybank,langmem2025}, or temporal decay with selective forgetting~\citep{gu2026fsfm}, their updates are driven by time or new inputs rather than memory usage itself. 
Another research approach uses reinforcement learning to estimate memory values~\citep{zhang2026memrl}, but this method only optimizes isolated entries without modifying the relational structure.
Evoked by memory reconsolidation, REALM instantiates a retrieval-driven lifecycle, dynamically reorganizing the activated local cognitive graph upon each retrieval.

\begin{figure*}[thbp]
    \centering
    \includegraphics[width=\linewidth]{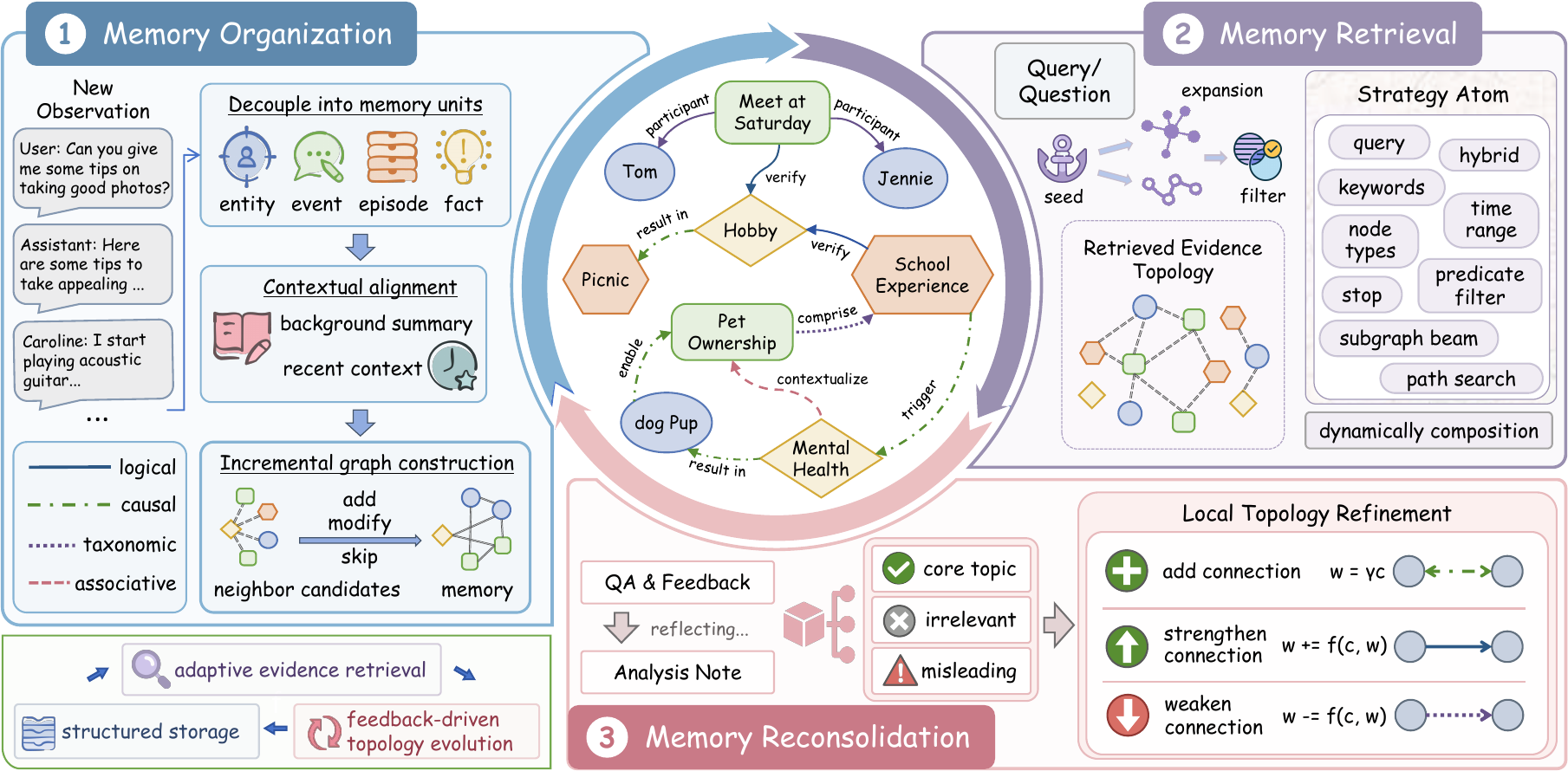}
    \caption{\textbf{Overview of REALM.} The framework organizes observations into an agentic cognitive graph memory, retrieves evidence through adaptive graph-search atoms, and reconsolidates the activated subgraph to evolve memory topology after use.}
    \label{fig:framework}
\end{figure*}

\section{Method}
\label{sec:method}

\subsection{Problem Formulation}
We model a long-horizon agent interacting with an environment over an information stream $\mathcal{O}=\{o_1,o_2,\cdots,o_T\}$.
Conventional agent memory paradigms sequentially evolve a memory graph $\mathcal{G}_t$ upon receiving $o_t$ via forward operations including insertion, consolidation, or forgetting, and passively extract a static subgraph $\mathcal{G}_q \subset \mathcal{G}_t$ to answer a query $q$.
In contrast, we contend that memory utilization yields crucial cognitive feedback that should actively drive structural refinement. We therefore introduce a post-retrieval memory reconsolidation phase. Given a query $q$ and its answer feedback $f$, the agent dynamically reorganizes the activated topology $\mathcal{G}_q$ into an optimized state $\mathcal{G}'_q$. The global memory state transition is thus formulated as:
\begin{equation}
\label{eq:recon-form}
\mathcal{G}_{t+1} = \text{Reconsolidating}(\mathcal{G}_t \setminus \mathcal{G}_q \cup \mathcal{G}'_q, \ f)
\end{equation}
The overall framework is illustrated in Figure~\ref{fig:framework}.

\subsection{Memory Organization: Unified Cognitive Graph}

\subsubsection{Graph Representation}
To enable schema-free local configurations, we model memory as a graph $\mathcal{G}=(\mathcal{V},\mathcal{E})$ where nodes correspond to different categories of memory units and edges encode their relations.  
Decomposing the structure into logically orthogonal dimensions, we formally define the node space as $\mathcal{V}\subset\{ \text{entity}, \text{event}, \text{episode}, \text{fact} \}$, and the edge space as $\mathcal{E} \subset \{ \text{logical}, \text{causal}, \text{hierarchical}, \text{associative} \}$. 
Formally, each agent-generated cognitive node $v\in\mathcal{V}$ is structured as:
\begin{equation}
v = \big( \kappa_v, \ c_v, \ des_v, \ k_v, \ t_v \big),
\end{equation}
where $\kappa_v$ denotes the node type, $c_v$ is the raw source content, $des_v$ represents the semantic description, $k_v$ signifies the type-specific keywords or names), and $t_v$ specifies the temporal span constraint if applicable. Correspondingly, each agent-generated directed edge $e_{ij} \in \mathcal{E}$ is formalized as:
\begin{equation}
e_{ij} = \big( v_i,\ v_j, \ \kappa_{ij}, \ des_{ij}, \ w_{ij} \big),
\end{equation}
where $\kappa_{ij}$ represents the relation type, $des_{ij}$ is the relational description text, and $w_{ij} \in [0, 1]$ denotes the confidence weight. This formulation decouples raw memory content from topology, providing the essential structural variables for downstream retrieval and topology evolution.

\subsubsection{Memory Construction}
Unlike conventional methods that separate insertion, merging, and forgetting, REALM unifies these operations into a single integration phase, treating forgetting as redundant~\citep{ong2025towards}. 
Upon receiving new observation $o_t$, the agent first extracts $m$ candidate memory units $X=\{x_i\}_{i=1}^{m}$ with the context of the current conversation summary $s_t$ and the recent information $c_t$, where each $x_i \in \mathcal{V}$ is autonomously typed by the agent.
Then for each candidate unit $x\in X$, the agent queries its related neighborhood nodes: $N(x)=N_{sim}(x)\cup N_{recent}(x)$, where $N_{sim}$ and $N_{recent}$ denote semantically similar and recently processed nodes, respectively. 
Conditioned on $N(x)$, the agent determines whether a new candidate should be \textit{added} to memory as a new node, \textit{merged} with an existing memory node, or \textit{skipped} and not saved in memory.
If $\text{add}$ or $\text{merge}$ is selected, the agent subsequently infers a set of directed relational edges $E_v$ between $x$ and $N(x)$ within the predefined edge space $\mathcal{E}$. 
This unified pipeline empowers the agent to autonomously determine not only \emph{whether to remember} an observation through selective retention, but also \emph{how to represent} it via dynamic topological integration.

\subsection{Memory Retrieval: Strategy Atom Combination}
Building on existing retrieval methods~\citet{peng2025graph}, REALM formalizes graph search into a dynamic pipeline of three sequential phases: seeding, expanding, and filtering. 
Rather than employing rigid heuristics, the agent autonomously formulates a composite retrieval policy $\pi$:
\begin{equation}
    \pi = \pi_{seed} \oplus \pi_{expand} \oplus \pi_{filter}
\end{equation}
based on the real-time search state, where each phase-specific policy combines atomic strategies from a predefined action space: 
\begin{equation}
    \pi_p = (a^p_1, a^p_2, \cdots, a^p_k), \quad a^p_i \in \mathcal{A}_p.
\end{equation}

\subsubsection{Seed Localization}

Given a query $q$, the agent first generates $\pi_{\text{seed}}$ consisting of multiple parameterized retrieval plans to identify starting nodes via keyword, query, or hybrid matching modes. 
Each plan is formalized as a tuple $p_i=(Q_i, K_i, T_i, \tau_i)$, explicitly defining the query string $Q_i$, keywords $K_i$, target node types $T_i$, and temporal constraints $\tau_i$. 
The aggregate initial seed set is formed via $\{S_i\} = \{\text{Retrieve}_{\text{seed}}(\mathcal{G}, p_i)\}$.

\subsubsection{Adaptive Evidence Expansion}

Starting from the seeds, the agent iteratively expands the evidence graph. At iteration $t$, it evaluates the collected memories $R_t$ (where $R_0 = \bigcup S_i$) to determine whether $R_t$ already provides enough evidence to answer the query. 
If sufficiency is met, search terminates; 
otherwise, it selects an active frontier $F_t \subseteq R_t$. For each frontier node $v \in F_t$, the LLM predicts an individualized expansion action:
\begin{equation}
a_v=(\text{mode}, \ \text{predicate},\ \text{decay},\ \text{inhibit})
\end{equation}
where components regulate the traversal behavior, edge-type predicates, temporal decay, and path inhibition, respectively. 
The newly discovered nodes are aggregated as 
\begin{equation}
    N_t = \bigcup_{v\in F_t}\text{Expand}(v,a_v),
\end{equation}
which update the state via $R_{t+1} = R_t \cup N_t$. 
During graph exploration, each newly discovered node $v_i \in N_t$ reached via edge $e_{ij}$ receives a context-aware access score $s_{\text{aces}}(v)$, formulated as:
\begin{equation}
s_{\text{aces}}(v) = \beta s_{\text{sim}}(q, \ des_{ij}) + (1-\beta) w_{ij} - \Delta_{decay}(t_v),
\end{equation}
where $s_{\text{sim}}$ measures the semantic similarity between query $q$ and edge description $des_{ij}$, $w_{ij}$ is the topological relation confidence, $\Delta_{devay}(t_v)$ scales the temporal penalty based on the permitted expansion decay and node's temporal span $t_v$, and $\beta$ is a balancing coefficient. Detailed strategy atoms and hyper-parameters are provided in the appendix. The expansion loop repeats until topological convergence or the predefined search depth and memory budgets are exhausted.

\subsubsection{Evidence Aggregation}
\label{evidence-aggregation}

Upon termination, the retrieved set $R$ is globally reranked. The final score for each node combines its accumulation weight and semantic relevance:
\begin{equation}
\text{Score}(v) = \alpha s_{\text{aces}}(v) + (1-\alpha)s_{\text{sim}}(v, q)
\end{equation}
The top-$K$ nodes constitute the final evidence set $\mathcal{V}_q = \text{Top}_K(R)$ for downstream generation. Crucially, $\mathcal{V}_q$ induces the activated local cognitive subgraph $\mathcal{G}_q=(\mathcal{V}_q, \mathcal{E}_q)$, where $\mathcal{E}_q=\{(u,v)\in\mathcal{E} \mid u,v\in \mathcal{V}_q\}$, which serves as the direct topology input for the subsequent memory reconsolidation.

\subsection{Memory Reconsolidation: Topology Evolution}
\label{memory-reconsolidation}
While memory organization structures the graph and retrieval utilizes it, reconsolidation leverages retrieval feedback to iteratively evolve the memory topology $\mathcal{G}$ beyond its initial semantic constraints into a usage-aware cognitive structure.

\subsubsection{Retrieval-Induced Structural Feedback}

For each interaction, the retrieval phase yields the activated subgraph $\mathcal{G}_q=(\mathcal{V}_q,\mathcal{E}_q)$ and task feedback $f$. 
Rather than modifying memory content, REALM exploits the topological co-utilization patterns within $\mathcal{V}_q$. The agent first distills the contextual guidance by inferring the underlying topic structure: 
\begin{equation}
    T_q=\textsc{InferTopicStructure}(q,\mathcal{V}_q),
\end{equation}
which summarizes the semantic relations among recalled memories to guide subsequent refinement.

\subsubsection{Local Topology Reconsolidation}

We model reconsolidation as an agent-driven topology decision process. Conditioned on $\mathcal{G}_q$ and $T_q$, the agent generates a modification set $\mathcal{D}=\{d_1,d_2,\ldots,d_n\}$, where each decision is parameterized as a 5-tuple $d=(v_i,v_j,a,r,c)$. Here, $v_i,v_j \in \mathcal{V}_q$ denote target nodes, $r$ is the relation type, $c \in [0,1]$ is the decision confidence, and $a \in \{\text{add}, \text{strengthen}, \text{weaken}\}$ defines the atomic edge operation space.
The agent may create new relations between previously disconnected memories when they are repeatedly activated under the same reasoning context. Existing relations can be strengthened when they consistently support successful retrieval, while weak or misleading relations can be suppressed when they repeatedly introduce irrelevant evidence.

\paragraph{Confidence-Guided Topology Update.}

To execute modifications without exceeding the valid edge weight boundaries, we unify the adaptive bounded adjustment. The confidence score determines the magnitude of topology modification. Let $w_{ij}$ be the current edge weight and $w_{\min}$ be the minimum allowable threshold. We first define a bounded representation $w_b=\max(w_{\min},\,\min(1,w_{ij}))$. The weight updating pipeline is then compactly formulated as:
\begin{equation}
w_{ij} \leftarrow \text{Clip}\Big( w_{ij} + \eta \cdot c, \ w_{\min}, \ 1 \Big),
\end{equation}
where
\begin{equation}
\eta =
\begin{cases}
0.8, & a=\text{add},\\ 
1-w_b, & a=\text{strengthen},\\
-(w_b-w_{\min}), & a=\text{weaken}.
\end{cases}
\end{equation}
This formulation smoothly regulates the update scale near the weight boundaries, protecting the topology from destabilization by isolated or duplicate retrieval instances.

\subsubsection{Closed-Loop Memory Evolution}

As formalized in Eq.\ref{eq:recon-form}, this mechanism establishes a seamless, closed-loop memory lifecycle: 
\begin{equation}
     \mathcal{G}_{t} \xrightarrow{\text{Retrieve}} (\mathcal{G}_q,f) \xrightarrow{\text{Reconsolidate}} \mathcal{G}_{t+1}.
\end{equation}
Through continuous task-driven interaction, the memory graph progressively adapts from an initially semantic organization into a usage-aware cognitive structure. 
Frequently co-utilized memory paths are dynamically reinforced while ineffective retrieval pathways are gradually weakened.
Therefore, long-term memory is no longer a static storage component, but an evolving cognitive graph.

\begin{table*}[thbp]
\centering
\small
\setlength{\tabcolsep}{6pt}
\begin{tabular}{l c c c c c}
\toprule
\multirow{2}{*}{\textbf{Method}} & \multicolumn{4}{c}{\textbf{LoCoMo-Category}} & \multirow{2}{*}{\textbf{Average}} \\
\cmidrule(lr){2-5}
& \textbf{Multi Hop} & \textbf{Temporal} & \textbf{Open Domain} & \textbf{Single Hop} & \\
\midrule
MIRIX~\citep{wang2025mirix}        & 54.26 & \underline{68.54} & 46.88 & 68.22 & 64.33 \\
Mem0~\citep{chhikara2025mem0}         & \underline{58.75} & 52.34 & 45.83 & 73.33 & 64.57 \\
Zep~\citep{rasmussen2025zep}          & 52.12 & 54.82 & 33.33 & 66.23 & 59.22 \\
MAGMA~\citep{jiang2026magma}       & 52.80 & 65.00 & \underline{51.70} & \underline{77.60} &  \underline{68.80}\\
Nemori~\citep{nan2025nemori}       & 56.90 & 64.90 & 48.50 & 76.40 & 68.70 \\
A-Mem~\citep{xu2026amem}              & 53.55 & 50.16 & 41.67 & 61.83 & 56.62 \\
\midrule
REALM (Ours) &\textbf{64.54} & \textbf{76.64} & \textbf{58.33} & \textbf{81.57} & \textbf{75.97} \\
\bottomrule
\end{tabular}
\caption{Accuracy (\%) on LoCoMo by question category using GPT-4o-mini as both the backbone model and the LLM judge. \textbf{Bold} and \underline{underlined} values indicate the best and second-best results respectively.}
\label{tab:locomo_results}
\end{table*}

\begin{table*}[thbp]
\centering
\small
\setlength{\tabcolsep}{4pt}
\resizebox{\columnwidth}{!}{
\begin{tabular}{l c c c c c c c}
\toprule
\multirow{2}{*}{\textbf{Method}} & \multicolumn{6}{c}{\textbf{LongMemEval-Category}} & \multirow{2}{*}{\textbf{Average}} \\
\cmidrule(lr){2-7}
& \textbf{\makecell[c]{single-session\\preference}} & \textbf{\makecell[c]{single-session\\assistant}} & \textbf{\makecell[c]{temporal\\reasoning}} & \textbf{\makecell[c]{multi-\\session}} & \textbf{\makecell[c]{knowledge\\update}} & \textbf{\makecell[c]{single-\\session user}}\\
\midrule
MIRIX & 53.30 & 63.60 & 25.60 & 30.10 & 52.60 & 72.90 & 43.49\\
Zep & 53.30 & 75.00 & \underline{54.10} & \underline{47.40} & \underline{74.40} & \textbf{92.90} & \underline{63.80}\\
MAGMA & \textbf{73.30} & \textbf{83.90} & 45.10 & \textbf{50.40} & 66.70 & 72.90 & 61.20\\
Nemori & \underline{62.70} & 73.20 & 43.00 & 51.40 & 52.60 & 77.70 & 56.20 \\
\midrule  
REALM (Ours) & 36.66 & \underline{82.14} & \textbf{56.69} & 46.28 & \textbf{88.89} & \underline{89.06} & \textbf{65.11}\\
\bottomrule
\end{tabular}
}
\caption{Accuracy (\%) on LongMemEval by question category using GPT-4o-mini as both the backbone model and the LLM judge. \textbf{Bold} and \underline{underlined} values indicate the best and second-best results respectively.}
\label{tab:longmemeval_results}
\vspace{-1em}
\end{table*}

\section{Experiment}
\label{sec:experiment}

To comprehensively evaluate the proposed framework, we conduct experiments from four complementary perspectives. Specifically, we aim to answer the following research questions:

\noindent\textbf{RQ1.} Does REALM consistently outperform existing long-term memory systems on representative long-term memory benchmarks?

\noindent\textbf{RQ2.} Does memory reconsolidation contribute to performance improvement, and is its effectiveness robust under different interaction orders?

\noindent\textbf{RQ3.} What enables REALM to remain effective prior to reconsolidation, and how does agentic autonomy shape memory organization and retrieval?

\noindent\textbf{RQ4.} How does memory reconsolidation continuously reshape memory organization and influence future retrieval behaviors?

We first introduce the experimental setup and then answer these research questions in the following sections.

\subsection{Experimental Setup}

\paragraph{Benchmarks \& Baselines.} 
We evaluate REALM on two standard long-term memory benchmarks: LoCoMo~\citep{maharana2024locomo} and LongMemEval\_S~\citep{wu2024longmemeval}. We compare our method against representative systems across three principal paradigms: (1) flat and temporal retrieval (Mem0, Zep); (2) multi-component and agentic adaptive memory (MIRIX, A-Mem, Nemori); and (3) strategy-guided graph architectures (MAGMA). Detailed benchmark statistics and baseline descriptions are deferred to the appendix.

\paragraph{Metrics.}
Following \citet{li2025memos}, we employ LLM-as-a-Judge \citep{zheng2023judging} for accuracy evaluation, excluding unanswerable queries. 
To ensure strict consistency, all baseline results from both newly conducted experiments and aligned literature utilize identical judge prompts and models, as detailed in the appendix.

\subsection{RQ1: Overall Performance}

Table~\ref{tab:locomo_results} and Table~\ref{tab:longmemeval_results} summarize the overall performance of REALM compared with representative long-term memory systems. 
The evaluation protocol and question scheduling strategy are provided in the appendix.
Our method consistently achieves the best performance across both benchmarks, demonstrating the effectiveness of the proposed framework in diverse long-term memory scenarios. 
On LoCoMo, REALM achieves an average score of 75.97, outperforming the strongest baseline by 7.17 points and ranking first across all four question categories. 
On LongMemEval, REALM also obtains the highest overall average (65.11), surpassing the previous best method by 1.31 points. 
The consistent improvements across two benchmarks with different evaluation settings indicate that the proposed framework generalizes well to a wide range of long-term memory tasks.

\begin{wrapfigure}{r}{0.5\columnwidth}
    \centering
    \includegraphics[width=\linewidth]{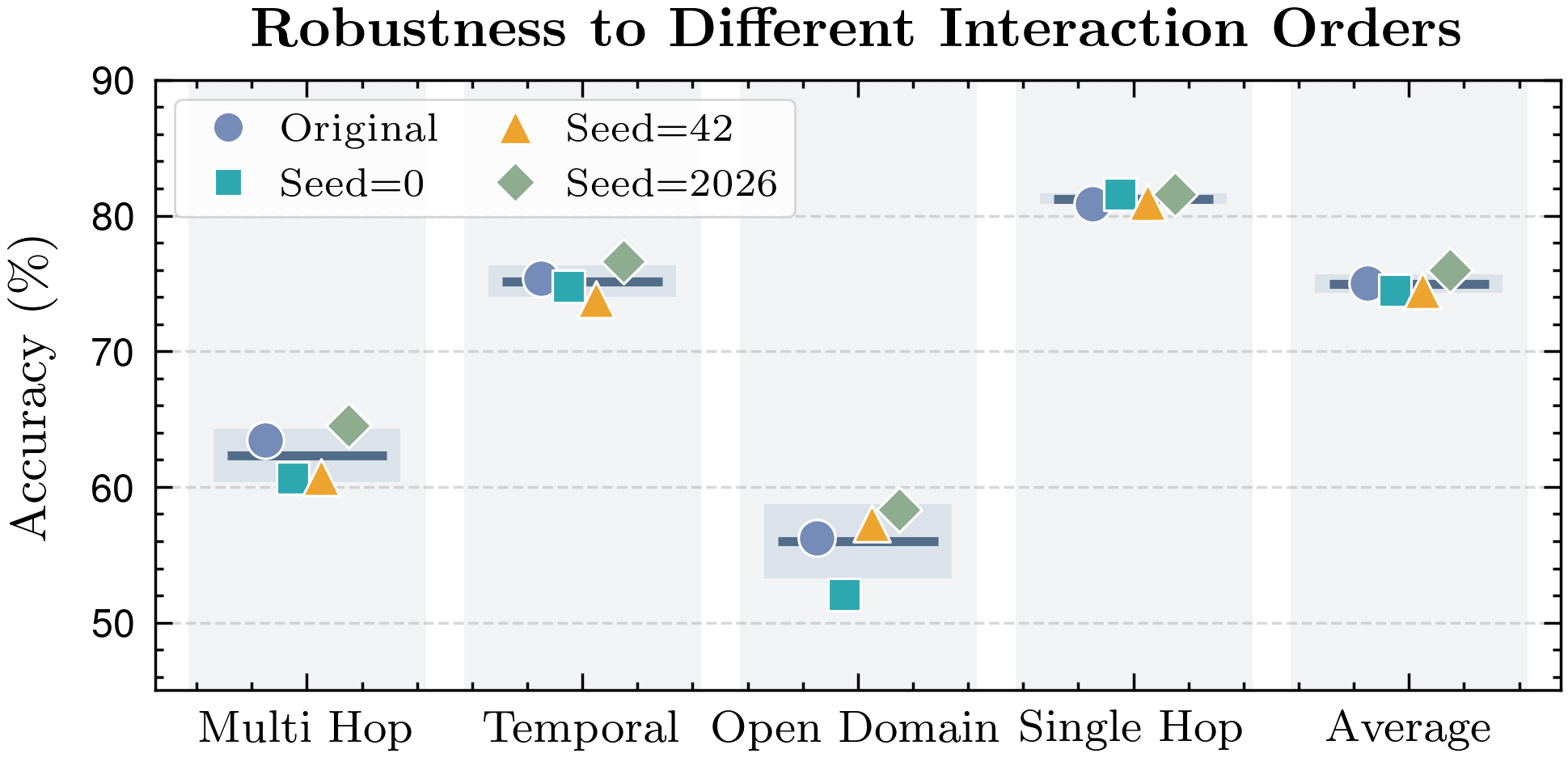}
    \caption{\textbf{Robustness to randomized question orders on LoCoMo.} Accuracy remains stable across original and shuffled orders, suggesting that reconsolidation improves memory topology rather than overfitting to a particular query sequence. The shaded region denotes performance variation.}
    \label{fig:stable_of_question_random_sequence}
\end{wrapfigure}


Looking into different question categories, the performance gains are particularly evident on memory-intensive reasoning tasks.
On LoCoMo, REALM improves multi-hop and temporal reasoning by more than 8 and 5 points, respectively, while also consistently outperforming existing methods on open domain and single-hop questions. 
These results indicate that the proposed framework is effective across both complex reasoning tasks requiring long-range memory integration and more straightforward retrieval scenarios.
A similar trend is observed on LongMemEval. REALM achieves the best performance on temporal reasoning and knowledge update, while remaining competitive on single-session assistant and single-session user. 
In particular, the improvement on knowledge update suggests that the framework effectively handles continuously evolving user information.
In contrast, the gains on temporal reasoning demonstrate its ability to leverage long-term interaction histories.
Overall, the results demonstrate that REALM provides a more effective long-term memory framework than existing approaches across diverse benchmarks and task types. 

\subsection{RQ2: Effect of Memory Reconsolidation}

\begin{table}[t]
\centering
\setlength{\tabcolsep}{5pt}
\newcommand{\posgain}[1]  
{\textcolor{green!50!black}{\scriptsize$_{+#1}$}}
\begin{tabular}{l cc cc}
\toprule
\multirow{2}{*}{\textbf{Category}} &
\multicolumn{2}{c}{\textbf{LoCoMo}} &
\multicolumn{2}{c}{\textbf{LongMemEval}} \\
\cmidrule(lr){2-3}\cmidrule(lr){4-5}
& \textbf{w/o recon} & \textbf{w/ recon}
& \textbf{w/o recon} & \textbf{w/ recon} \\
\midrule
Multiple    & 59.57 & 64.54\posgain{4.97} & 43.80 & 46.28\posgain{2.48} \\
Temporal    & 74.14 & 76.64\posgain{2.50} & 55.12 & 56.69\posgain{1.57} \\
Single(-P)      & 81.09 & 81.57\posgain{0.48} & 30.00 & 36.66\posgain{6.66} \\
Open/Update & 53.12 & 58.33\posgain{5.21} & 84.72 & 88.89\posgain{4.17} \\
\midrule
Average & 73.96 & 75.97\posgain{2.01} & 62.98 & 65.11\posgain{2.13} \\
\bottomrule
\end{tabular}
\caption{Effect of reconsolidation across different categories on LoCoMo and LongMemEval.  Subscripts report absolute gains from enabling reconsolidation; ``Single(-P)'' denotes the single-session preference category in LongMemEval because the other single-session categories remain unchanged.}
\label{tab:ablation-category}
\vspace{-1em}
\end{table}


Memory reconsolidation is the key mechanism distinguishing REALM from existing long-term memory systems. 
To evaluate its effectiveness, we conduct two ablation studies: (1) removing the reconsolidation module while keeping all other components unchanged, and (2) randomizing the question order on LoCoMo to test whether the gains depend on a particular interaction sequence.

\paragraph{Effect of Memory Reconsolidation.}
Table~\ref {tab:ablation-category} shows that removing memory reconsolidation consistently degrades performance on both benchmarks. On LoCoMo, enabling reconsolidation improves the overall accuracy by 2.01 points, with the largest gains observed on Multi-Hop (+4.97) and Open-Domain (+5.21) questions. Similar improvements are observed on LongMemEval, where reconsolidation increases the overall performance by 2.13 points, particularly benefiting Single-session Preference (+6.66) and Knowledge Update (+4.17). These improvements indicate that updating the cognitive graph after each retrieval enables the system to progressively strengthen useful memory associations and improve subsequent retrieval quality.

\paragraph{Robustness to Interaction Order.}
One possible concern is that the gains from reconsolidation may arise from adapting to a specific question sequence. To verify this, we evaluate REALM under multiple randomly shuffled question orders on LoCoMo. As shown in Figure~\ref{fig:stable_of_question_random_sequence}, the performance across different random seeds remains highly consistent for all question categories, with only minor fluctuations around the mean. This result demonstrates that the effectiveness of memory reconsolidation is largely independent of the interaction order. Rather than memorizing a particular sequence of queries, reconsolidation continuously refines the memory graph according to the retrieval process itself, leading to stable improvements under different evaluation orders.


\subsection{RQ3: Dissecting Agentic Memory Autonomy}


\begin{figure}[thbp]
    \centering
    \includegraphics[width=0.8\linewidth]{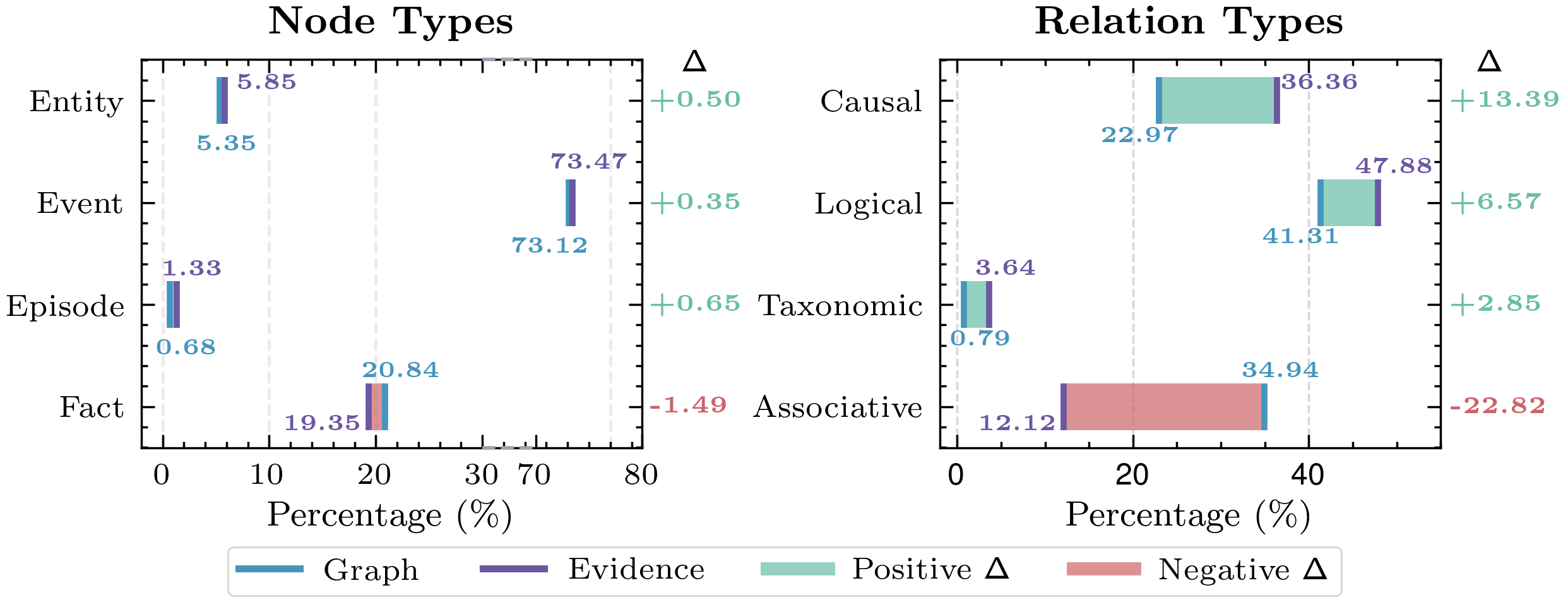}
     \caption{\textbf{Deviation between graph composition and evidence retrieval.} The left figure illustrates that heterogeneous memory node types remain unbiased in both graph structure and evidence utilization, while the right figure shows that different relationship types are selectively exploited according to downstream reasoning requirements.
    }
    \label{fig:type_distribution}
    \vspace{-1em}
\end{figure}




This section investigates why REALM remains competitive even without reconsolidation by analyzing memory organization, utilization, and retrieval strategy selection.

\paragraph{Memory Organization and Utilization.}

We first compare the distributions of node and relation types in the cognitive graph with those in the retrieved evidence, as shown in Figure~\ref{fig:type_distribution}. Node distributions are highly consistent, with differences below 1.5 percentage points across all types. This suggests that each memory type contributes to retrieval according to both its semantic role and its frequency in the graph. Even less frequent types, such as episode nodes, are retrieved when needed rather than overlooked, indicating that the automatically constructed graph preserves diverse memory forms without introducing retrieval bias. 
Relation usage exhibits a much stronger task-dependent pattern. During retrieval, causal and logical relations increase by 13.39 and 6.57 percentage points, respectively, whereas associative relations become markedly less frequent. This suggests that different relation types naturally play distinct functional roles: associative relations mainly maintain broad semantic connectivity, while causal and logical relations form the primary reasoning paths for question answering. Overall, these results show that the heterogeneous cognitive graph is not only automatically organized by the agent but also effectively exploited to support downstream reasoning.

\begin{wrapfigure}{r}{0.5\columnwidth}
    \centering
    \includegraphics[width=1\linewidth]{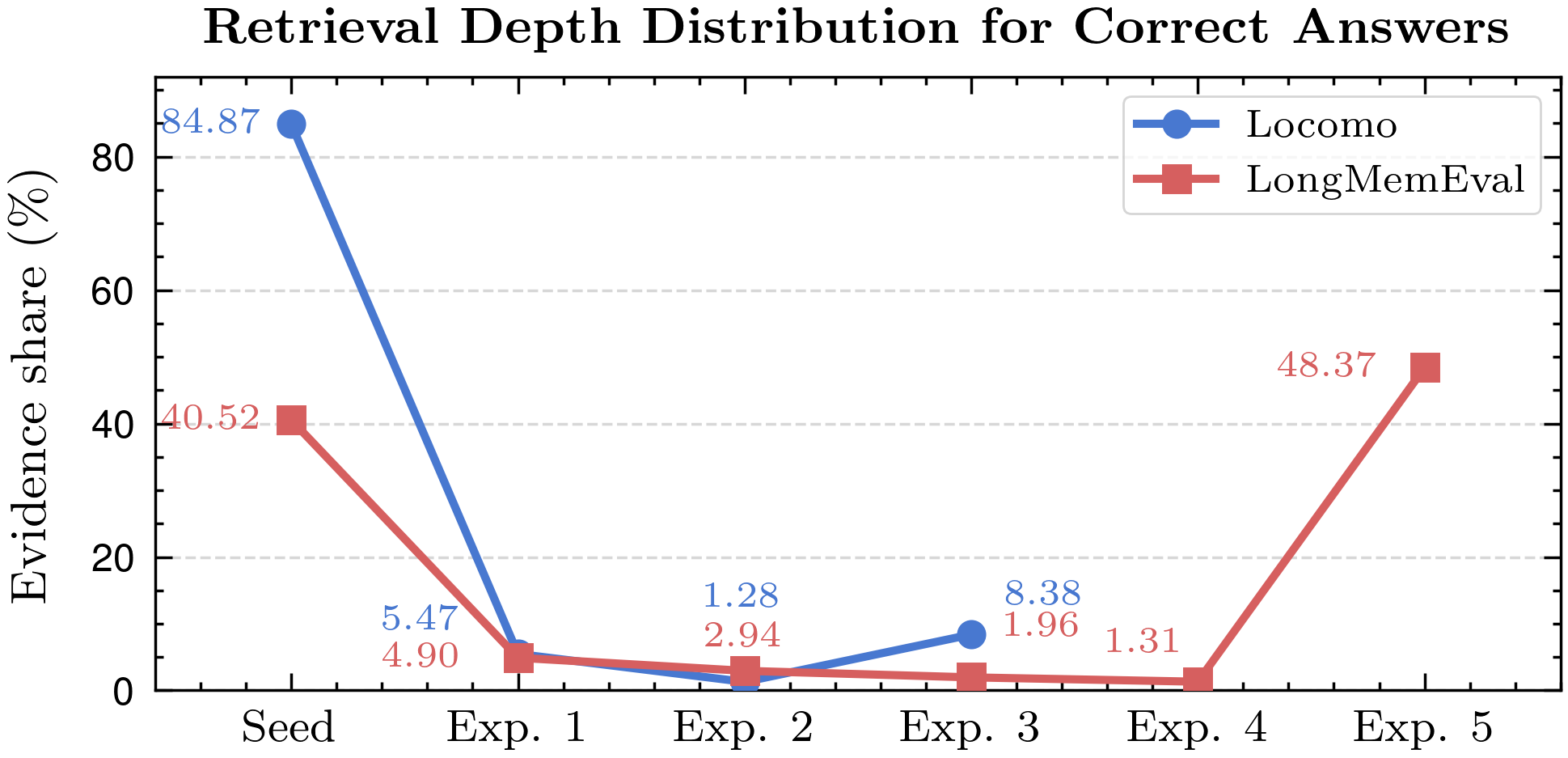}
    \caption{\textbf{Retrieval depth for correct answers on LoCoMo and LongMemEval.} The proportion first decreases and then increases with expansion depth, indicating that straightforward queries are answered early, and more challenging ones require deeper graph exploration.}
    \label{fig:evidence_layer_percentage}
\end{wrapfigure}

\paragraph{Retrieval Efficiency.}
We further analyze the expansion depth at which supporting evidence is first retrieved. As shown in Figure~\ref{fig:evidence_layer_percentage}, 84.87\% of LoCoMo evidence is found immediately from the seed nodes, found directly from the seed nodes, with only a small fraction requiring one to three additional expansion steps. On LongMemEval, 40.52\% of evidence is retrieved directly from the seed layer, whereas a considerable proportion is found only after reaching the maximum expansion depth. 
This pattern reflects two complementary properties of the proposed retrieval mechanism. For relatively straightforward queries, relevant memories are organized close to effective retrieval entry points, enabling efficient retrieval with minimal graph exploration. For more challenging questions, the cognitive graph still provides sufficient structural connectivity for the retrieval agent to progressively discover distant evidence through multi-hop expansion, indicating both retrieval efficiency and scalability.




\begin{table}[thbp]
\centering
\begin{tabular}{l c c}
\toprule
\textbf{Strategy} & \textbf{Acc (\%)} & \textbf{Correct (\#)} \\
\midrule
path\_search only & 39.84 & 147 \\
subgraph\_beam only & 43.09 & 159 \\
adaptive selection & \textbf{43.36} & \textbf{160} \\
\bottomrule
\end{tabular}
\caption{Effect of expansion strategy selection on questions requiring graph expansion. Adaptive selection slightly improves over fixed path search and subgraph beam, showing the value and current limitation of strategy composition.}
\label{tab:adaptive_strategy_selection}
\vspace{-1em}
\end{table}

\paragraph{Adaptive Retrieval Strategy.}
Finally, we evaluate whether allowing the agent to compose retrieval strategies adaptively is beneficial. Table~\ref{tab:adaptive_strategy_selection} shows that adaptive strategy selection achieves the highest accuracy, slightly outperforming either fixed expansion strategy. 
Although the improvement over always using ``subgraph\_beam'' is modest, this result is consistent with the observed strategy distribution, where the agent selects subgraph-based expansion for most queries while switching to alternative traversal strategies only when necessary. 
This behavior suggests that adaptive composition does not seek diversity for its own sake; instead, it learns to recover the strongest default strategy while preserving the flexibility to handle structurally different retrieval scenarios. 
At the same time, the relatively small margin also reveals a limitation of the current design: the benefit of adaptive retrieval ultimately depends on reliable strategy selection, and inaccurate decisions may reduce the advantage over carefully designed fixed strategies.
These analyses suggest that agents can also improve by autonomously constructing memory that aligns well with retrieval demands.




\subsection{RQ4: Memory Evolution through Reconsolidation}

\begin{wrapfigure}{r}{0.5\columnwidth}
    \centering
    \includegraphics[width=\linewidth]{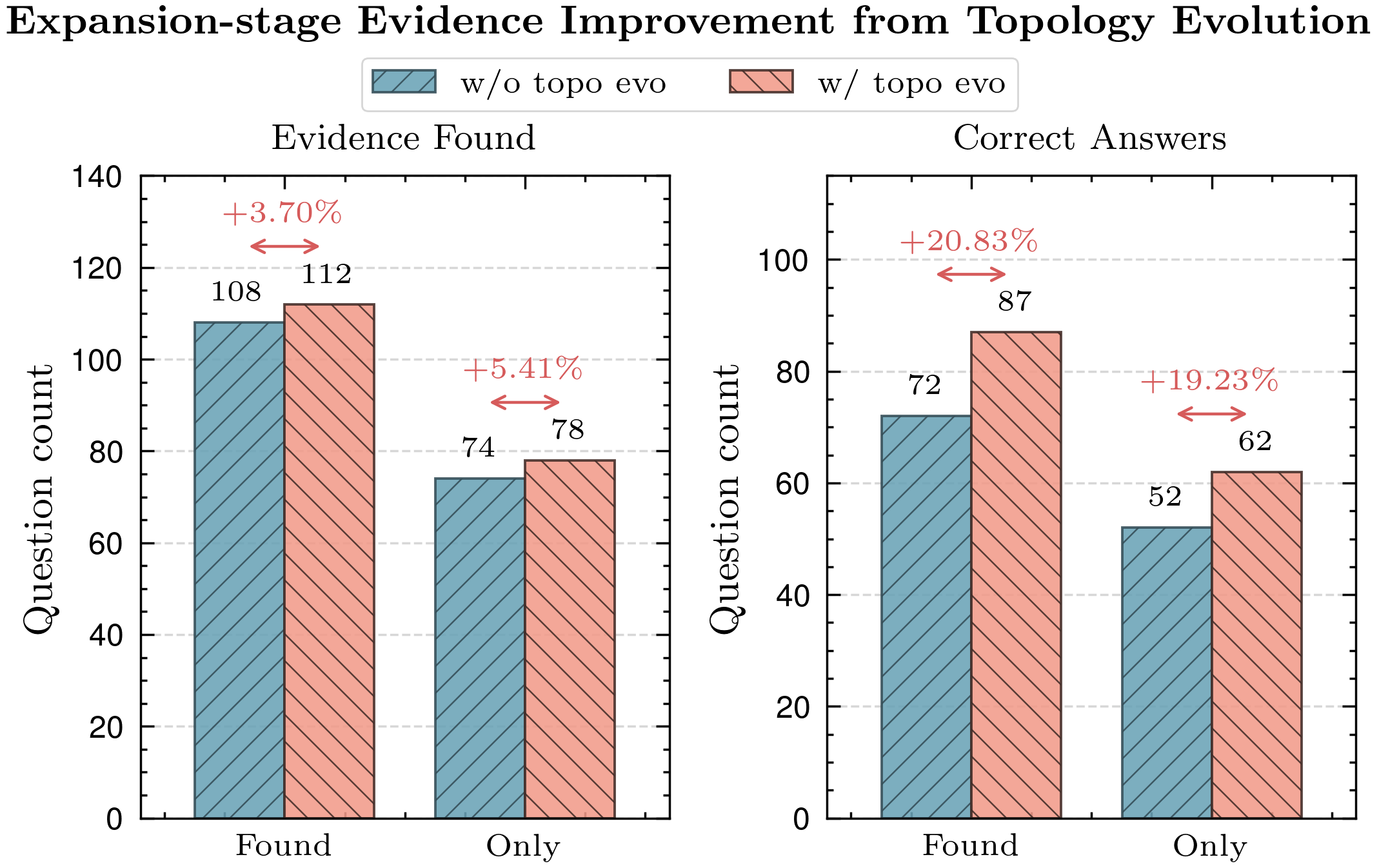}
    \caption{\textbf{Evidence gains from topology evolution during the expansion stage.} ``Found'' denotes cases where the supporting evidence is retrieved during expansion, and ``Only'' those where it is retrieved exclusively during expansion.}
    \label{fig:topology_evolution_effect}
\end{wrapfigure}





While the previous section analyzes the static memory organization and retrieval mechanism, this section investigates how memory reconsolidation continuously improves future retrieval through graph evolution. We compare retrieval behaviors before and after topology evolution to understand how reconsolidation influences evidence discovery and utilization over continual interactions.

As shown in Figure~\ref{fig:topology_evolution_effect}, after enabling memory reconsolidation, more questions successfully retrieve supporting evidence during graph expansion, resulting in a corresponding increase in correctly answered questions. More importantly, topology evolution brings substantially larger improvements in evidence utilization than in evidence discovery. The proportion of questions that successfully retrieve supporting evidence increases by only 3.70\% and 5.41\%, whereas the proportion of correctly answered questions after evidence retrieval increases by 20.83\% and 19.23\%, respectively. This indicates that memory reconsolidation does not primarily improve performance by retrieving substantially more memories. Instead, it continuously reorganizes the cognitive graph so that retrieved evidence becomes more relevant and informative for downstream reasoning.

These results suggest that the value of memory reconsolidation lies not in expanding the retrievable memory space, but in continuously improving the quality of memory organization through retrieval-driven graph evolution. As interactions accumulate, the cognitive graph becomes increasingly aligned with future retrieval requirements, thereby complementing static memory organization.

\section{Conclusion}

This work conceptualizes LLM agent long-term memory as a dynamic, closed-loop lifecycle, demonstrating that memory networks can actively adapt through utilization rather than merely expanding with new inputs. By treating retrieval not as a passive terminal checkpoint but as a continuous feedback loop, our framework, REALM, instantiates this lifecycle over a heterogeneous cognitive graph, enabling post-retrieval local topology updates driven by co-utilization patterns. 
Empirical evaluations indicate that introducing retrieval-driven memory reconsolidation effectively mitigates the rigidity of conventional predefined memory architectures, yielding robust performance gains across long-term memory scenarios. Further analyses reveal that this topology evolution induces highly compact evidence clustering, successfully organizing scattering interactions into structural subgraphs that facilitate collective recall. We hope that shifting from a flat accumulation model to a continuous lifecycle perspective will inspire future research into adaptive, self-evolving memory foundations for intelligent agents.

\bibliography{main}
\bibliographystyle{rlc}

\appendix

\section{Experimental Details}

\subsection{Benchmarks}
We evaluated REALM on two commonly used long-term memory benchmarks.
\begin{itemize}[leftmargin=*]
    \item \textbf{LoCoMo~\citep{maharana2024locomo}.} LoCoMo evaluates LLM-based agents over extended interaction histories. Each sample contains multi-session conversations and question-answering tasks that require locating, integrating, and reasoning over information distributed across the dialogue history. 
    \item \textbf{LongMemEval~\citep{wu2024longmemeval}.} LongMemEval focuses on long-context and long-term memory abilities of LLM-based agents. In our experiments, we tested only LongMemEval\_S, which require memory-intensive question answering and covers various question types.
\end{itemize}

\subsection{Baselines}
We compare our method against representative long-term memory systems covering the major design paradigms in agent memory. Mem0~\citep{chhikara2025mem0} represents the classical paradigm of memory extraction and management with flat retrieval. MIRIX~\citep{wang2025mirix} decomposes memory into multiple components, while Zep~\citep{rasmussen2025zep} introduces temporal memory structures. A-Mem~\citep{xu2026amem} dynamically determines memory formation through agentic decisions. Nemori~\citep{nan2025nemori} further models memory evolution through adaptive memory granularity and predictive retrieval. MAGMA~\citep{jiang2026magma} integrates a multi-graph memory architecture with strategy-guided graph traversal to enhance long-term memory. Together, these methods cover the principal design paradigms of modern long-term agent memory systems, providing a comprehensive set of baselines for comparison. The following is a detailed description of each baseline method.
\begin{itemize}[leftmargin=*]
    \item \textbf{Mem0~\citep{chhikara2025mem0}.} Mem0 maintains a flat memory store through an extraction-update pipeline: an LLM first extracts salient facts from each new message pair conditioned on a running conversation summary and recent history, then a second LLM call retrieves the top-$s$ semantically similar existing memories and issues one of four tool-call operations (\texttt{ADD}, \texttt{UPDATE}, \texttt{DELETE}, \texttt{NOOP}) to reconcile the new fact with the existing store, without modeling relations between memories.

    \item \textbf{MIRIX~\citep{wang2025mirix}.} MIRIX decomposes memory into six predefined components (Core, Episodic, Semantic, Procedural, Resource, and Knowledge Vault), each maintained by a dedicated memory manager agent coordinated by a meta memory manager that routes incoming information to the relevant components. At inference time, an active retrieval mechanism first infers a topic from the ongoing context and then retrieves the top-$k$ entries from each memory component independently.

    \item \textbf{Zep~\citep{rasmussen2025zep}.} Zep organizes memory as a temporally-aware knowledge graph (Graphiti) comprising three hierarchical subgraphs: episodes (raw messages), semantic entities and facts extracted from episodes, and higher-level community summaries obtained via label propagation. Facts carry bi-temporal validity intervals, and newly ingested edges can invalidate contradicting existing edges. Retrieval combines cosine similarity, BM25, and breadth-first graph search, followed by reranking.

    \item \textbf{A-Mem~\citep{xu2026amem}.} Following the Zettelkasten method, A-Mem represents each interaction as an atomic note enriched with LLM-generated keywords, tags, and a contextual description. When a new note is added, the system retrieves its top-$k$ nearest notes by embedding similarity and prompts an LLM to decide which links to establish among them; the same neighborhood is then passed to a separate memory evolution step, where an LLM updates each neighbor's context, keywords, and tags in light of the new note, allowing existing memories to evolve as new experience arrives.

    \item \textbf{Nemori~\citep{nan2025nemori}.} Nemori separates memory construction into episodic integration and semantic distillation. Raw messages are first partitioned into coherent episodes and rewritten into narrative form; each episode is then compared against an LLM-synthesized anticipatory schema retrieved from existing knowledge, and only the prediction-error residual, i.e., information the schema fails to anticipate, is distilled into semantic memory, following the predictive-coding principle that predictable content is redundant.

    \item \textbf{MAGMA~\citep{jiang2026magma}.} MAGMA represents each memory event across four orthogonal relation graphs (semantic, temporal, causal, and entity). Retrieval is formulated as policy-guided graph traversal: a router classifies query intent (\texttt{WHY}/\texttt{WHEN}/\texttt{ENTITY}) to reweight edge types, fuses anchor nodes across embedding, keyword, and temporal signals via reciprocal rank fusion, and performs an intent-weighted beam search before linearizing the retrieved subgraph into context. Memory evolution follows a dual-stream design that decouples fast temporal-edge ingestion from asynchronous LLM-based causal/entity edge consolidation.
\end{itemize}

\subsection{Memory Graph Configuration.} 
During retrieval, the number of seed plans for LoCoMo is limited to 1, with a maximum expansion depth of 3; for LongMemEval, the number of seed plans is limited to 3, with a maximum expansion depth of 5; and the maximum number of retrieved nodes is 10. The $\alpha$ value for evidence aggregation is set to 0.8, and the $\beta$ value for access score is set to 0.6.
After each question is evaluated, memory reconsolidation is performed once, and the memory graph is updated. Each question is answered only once, simulating the continuous evolution of memory in real-world usage scenarios.
Our framework does not require specialized hardware and can be executed on standard computing environments. All experiments rely on API-based LLM inference and lightweight graph operations.

\subsection{Evaluation Protocol and Question Scheduling}
\label{app:evaluation_protocol}
The two benchmarks LoCoMo~\citep{maharana2024locomo} and LongMemEval~\citep{wu2024longmemeval} adopt different evaluation protocols due to their distinct data characteristics.
Each LoCoMo sample contains a long conversation together with multiple questions of different types. 
Accordingly, we evaluate the questions sequentially within each sample. 
After answering each question, the memory graph undergoes one round of memory reconsolidation, allowing its structure to evolve throughout the evaluation process.
In contrast, each LongMemEval sample contains only a single question of a specific type, which does not provide sufficient opportunities for memory reconsolidation.
Therefore, for each sample, we use GPT-4o-mini and GPT-4.1 to generate one to six additional questions based on the corresponding oracle session, simulating historical memory usage before evaluating the original benchmark question.
Following prior agent memory studies~\citep{li2025memos,nan2025nemori}, we exclude unanswerable questions from both benchmarks. We choose LoCoMo and LongMemEval\_S because they are the two most widely used benchmarks for evaluating long-term memory in LLM-based agents. Our goal is not to optimize leaderboard performance, but to validate the effectiveness of memory reconsolidation as a continual memory organization mechanism.

\subsection{Evaluation Details}
Due to the limited availability of reproducible implementations and the substantial computational cost of reproducing all baselines, we reuse officially reported results whenever possible.
All compared methods are evaluated under the same backbone and evaluation protocol whenever possible. Specifically, we only compare methods using ``gpt-4o-mini'' as the backbone model and adopt the same ``gpt-4o-mini''-based LLM-as-a-Judge prompts as used in the corresponding papers. For methods whose official implementations or evaluation results are unavailable under our setting, we reproduce them ourselves. In the main results, the performance of MIRIX and Mem0 is taken from the MemOS paper~\citep{li2025memos}, while the results of MAGMA and Nemori are adopted from the MAGMA paper~\citep{jiang2026magma}. We reproduced the A-Mem results under the same experimental settings. 

\begin{algorithm}[tb]
\caption{Cognitive Graph Memory Organization}
\label{alg:memory_organization}
\textbf{Input}: Observation $o_t$, Conversation Summary $s_t$, Recent Updates $c_t$, Memory Graph $\mathcal{G}$\\
\textbf{Output}: Updated Memory Graph $\mathcal{G}$
\begin{algorithmic}[1]
\STATE $X \leftarrow \textsc{ExtractMemoryUnits}(o_t, s_t, c_t)$
\FOR{$x$ in $X$}
    \STATE $N \leftarrow \textsc{RetrieveSimilarAndRecent}(x)$
    \STATE $a \leftarrow \textsc{PredictOperation}(x, N)$
    \STATE $v \leftarrow \textsc{Execute}(a)$
    \STATE $E \leftarrow \textsc{InferRelations}(v, N)$
    \STATE \textsc{UpdateGraph}$(E)$
\ENDFOR
\STATE \textbf{return} $\mathcal{G}$
\end{algorithmic}
\end{algorithm}

\begin{algorithm}[tb]
\caption{Retrieval via Strategy Atom Combination}
\label{alg:memory_retrieval}
\textbf{Input}: Query $q$, Memory Graph $\mathcal{G}$\\
\textbf{Output}: Evidence Set $\mathcal{V}_q$, Activated Subgraph $\mathcal{G}_q$
\begin{algorithmic}[1]
\STATE $\pi_{seed} \leftarrow \textsc{ComposeSeedStrategy}(q)$
\STATE $R \leftarrow \emptyset$
\FOR{each retrieval plan $p \in \pi_{seed}$}
    \STATE $R \leftarrow R \cup \textsc{SeedRetrieve}(\mathcal{G},p)$
\ENDFOR
\WHILE{\textbf{not} \textsc{StopRetrieval}$(R)$}
    \STATE $F \leftarrow \textsc{SelectFrontier}(R)$
    \FOR{each node $v\in F$}
        \STATE $a_v \leftarrow \textsc{ComposeExpandStrategy}(v,R,q)$
        \STATE $N \leftarrow \textsc{Expand}(v,a_v)$
        \STATE \textsc{UpdateAccessScore}$(N)$
        \STATE $R \leftarrow R\cup N$
    \ENDFOR
\ENDWHILE
\STATE \textsc{Rerank}$(R,q)$
\STATE $\mathcal{V}_q \leftarrow \textsc{TopK}(R)$
\STATE $\mathcal{G}_q \leftarrow \textsc{InduceSubgraph}(\mathcal{G},\mathcal{V}_q)$
\STATE \textbf{return} $\mathcal{V}_q,\mathcal{G}_q$
\end{algorithmic}
\end{algorithm}

\begin{algorithm}[tbhp]
\caption{Feedback-Driven Topology Evolution}
\label{alg:topology_evolution}
\textbf{Input:} Question $q$, Retrieved Subgraph $\mathcal{G}_q$, Feedback $f$, Memory Graph $\mathcal{G}$\\
\textbf{Output:} Updated Graph $\mathcal{G}$
\begin{algorithmic}[1]
\STATE $T \leftarrow \textsc{InferTopicStructure}(q,\mathcal{G}_q)$
\STATE $P^{key}, P^{noise} \leftarrow \textsc{IdentifyGroup}(q,\mathcal{G}_q, f, T)$
\STATE $\mathcal{D} \leftarrow \textsc{AgentEdit}(T,P,\mathcal{G}_q))$
\FORALL{$d=(v_i,v_j,a,r,c)\in\mathcal{D}$}
    \IF{$a=\texttt{create}$}
        \STATE \textsc{AddEdge}$(v_i,r,v_j,\eta c)$
    \ELSE
        \IF{$a=\texttt{strengthen}$}
            \STATE $w_{ij} \leftarrow w_{ij} + \eta c(1-w_{ij})$
        \ELSIF{$a=\texttt{weaken}$}
            \STATE $w_{ij} \leftarrow w_{ij} - \eta c w_{ij}$
        \ENDIF
    \ENDIF
\ENDFOR
\STATE \textbf{return} $\mathcal{G}$
\end{algorithmic}
\end{algorithm}

\section{Algorithms for REALM}
To facilitate understanding and reproducibility, we provide the complete pseudo-code for the proposed REALM framework. The algorithms correspond to the three stages introduced in the method section: memory organization, memory retrieval, and memory reconsolidation. They summarize the control flow and interactions among the LLM, memory graph, and retrieval modules while omitting implementation-specific details for readability. The pseudo-code is presented in Algorithms~\ref{alg:memory_organization}, \ref{alg:memory_retrieval}, and~\ref{alg:topology_evolution}, respectively.



\section{Retrieval Details}

\subsection{Strategy Atoms}
\label{app:strategy_atoms}
This section walks through the three retrieval prompts (Prompt~\ref{prompt:seed_planner_prompt}, \ref{prompt:retrieval_sufficiency_prompt}, \ref{prompt:expansion_prompt}) by option, describing how each option that actively affects retrieval is executed.

\paragraph{Seed Retrieval Planner (Prompt~\ref{prompt:seed_planner_prompt}).}
Each seed retrieval plan $p_i=(Q_i,K_i,T_i,\tau_i)$ involves three key decisions: matching mode that determines how $Q_i$ and $K_i$ are used to score candidate nodes, $T_i$ and $\tau_i$ are used to filter node types and time, respectively. Multiple seed plans issued for one question are executed independently and merged into a single seed set by node id.

\begin{itemize}[leftmargin=*]
\item \textbf{matching\_mode} is determines how $Q_i$ and $K_i$ are used to score candidate nodes:
\begin{itemize}[leftmargin=*]
    \item \texttt{query\_match}. Only $Q_i$ is embedded and matched against node embeddings by cosine similarity; keywords are not used.
    \item \texttt{keywords\_match}. Only $K_i$ is matched against nodes through lexical, inverted-index keyword search; the query embedding is not used.
    \item \texttt{hybrid\_match}. Both signals are computed for every node and combined as a weighted sum (embedding weight 0.6, keyword weight 0.4 by default), so a node found by either signal can be recalled.
\end{itemize}
\item \textbf{allowed\_node\_types} ($T_i$) is applied as a memory node type filter in all three modes, but is not strictly enforced. If restricting to $T_i$ together with the score threshold would leave no candidates, the filter is relaxed in two steps: first the score threshold is dropped while keeping $T_i$, and then if the result is still empty, $T_i$ itself is dropped. Therefore, seed retrieval never returns an empty set.
\item\textbf{time\_range} ($\tau_i$) is only enforced when matching mode is \texttt{hybrid\_match}: a node is kept only if its own time interval overlaps $\tau_i$, nodes outside this window are excluded outright, and a node carrying no resolvable timestamp is treated as always inside the window and kept by default.
\end{itemize}

\paragraph{Retrieval Sufficiency Controller (Prompt~\ref{prompt:retrieval_sufficiency_prompt}).}
Before each expansion round, this atom inspects the current retrieved nodes set and returns two fields that control the loop:
\begin{itemize}[leftmargin=*]
    \item \textbf{is\_enough}. If true, retrieval terminates immediately with the current retrieved memory nodes.
    \item \textbf{nodes\_to\_expand}. If \texttt{is\_enough} is false, this list of node ids becomes the next expansion frontier $F_t$. If the list is empty or none of its ids belong to the current nodes set, the frontier instead defaults to the highest-scoring nodes that have not yet been expanded.
\end{itemize}

\paragraph{Graph Expansion Policy (Prompt~\ref{prompt:expansion_prompt}).} For each frontier node $v\in F_t$, this atom produces an expansion action $a_v=(\text{mode},\ \text{predicate},\ \text{decay},\ \text{inhibit})$ together with \texttt{score\_threshold} and \texttt{reasoning}. Every candidate edge $e$ reachable from $v$, with endpoint node $u$, is scored as
\begin{equation}
    score(e,u) = w_{sem}\cdot\text{sim}(e,r) + w_{edge}\cdot\omega_e + \delta_{temporal},
\end{equation}
clipped to $[0,1]$, where $\text{sim}(e,r)$ is the cosine similarity between the embedding of edge $e$'s description and the embedding of the \textbf{reasoning} text $r$, $\omega_e$ is the current edge weight, $\delta_{temporal}$ is the penalty contributed by \textbf{decay} (zero when disabled), and $w_{sem}=0.6,w_{edge}=0.4$ by default. A candidate is discarded before scoring if it revisits a node already on the current path, has already been reached from another seed, or is excluded by \textbf{predicate}; after scoring, it is further discarded if its score falls below \texttt{score\_threshold}.
\begin{itemize}[leftmargin=*]
    \item \textbf{predicate} (\texttt{layer\_predicate\_include}): restricts candidate edges to a given relation layer sets from \{logical, causal, taxonomic, associative\}, optionally one predicate within that layer, and a traversal direction (forward, backward, or both); this is a hard filter applied before scoring.
    \item \textbf{decay} (\texttt{temporal\_decay}): is used to compare a candidate node's timestamp against a reference window (time $\pm$ $\Delta t$). \texttt{hard} mode drops out-of-window candidates entirely; \texttt{soft} mode instead applies a continuous penalty that grows with the distance beyond the window boundary. Nodes lacking timestamp information are kept with a fixed penalty under \texttt{missing\_time\_mode=keep}, or dropped outright under \texttt{missing\_time\_mode=drop}.
    \item \textbf{inhibit} (\texttt{conflict\_inhibit}): is used to discard any candidate edge whose predicate is \texttt{contradicts} before scoring when enabled.
\end{itemize}
The \textbf{mode} atom determines how candidates in the expansion stage are selected and carried into the next hop:
\begin{itemize}[leftmargin=*]
    \item \texttt{subgraph\_beam}. At each hop, aggregate all candidate nodes from the active frontiers, remove duplicates based on the target node ID (retaining the entry with the highest score), and admit them in order of score until a single node budget shared by the whole seed (\texttt{subgraph\_max\_nodes}) is exhausted. The frontier is then completely replaced by the newly accepted nodes.
    
    \item \texttt{path\_search}. Each active path is extended independently. The generated paths are deduplicated based on their terminal nodes (retaining the path with the highest cumulative path score, calculated as the average hop score minus a fixed penalty of $0.03$ for each additional hop), and only the top-\texttt{path\_top\_k} paths remain active for the next hop, for a maximum of \texttt{path\_max\_depth} hops. Paths that are pruned at a particular hop will not be reconsidered.
    
\end{itemize}

\section{Prompts}

For completeness, we provide the detailed prompt templates across all three functional stages of our framework:
\begin{itemize}
    \item Memory Organization:
    \begin{itemize}
        \item Memory Unit Construction (Prompt~\ref{prompt:memory_unit_construction_prompt}): Extracts memory units and updates dialogue summaries.
        \item Memory Operation Selection (Prompt~\ref{prompt:memory_construction_operation_prompt}): Decides node insertion, merging, or skipping.
        \item Memory Relation Prediction (Prompt~\ref{prompt:memory_relation_decision_prompt}): Determines edge operations and semantic relations between nodes.
    \end{itemize}
    \item Retrieval Stage:
        \begin{itemize}
            \item Seed Planning (Prompt~\ref{prompt:seed_planner_prompt}): Selects retrieval modes and constraints to identify anchor nodes.
            \item Sufficiency Assessment (Prompt~\ref{prompt:retrieval_sufficiency_prompt}): Evaluates evidence completeness and identifies candidate expansion nodes.
            \item Retrieval Expansion (Prompt~\ref{prompt:expansion_prompt}): Defines traversal modes and edge constraints for graph propagation.
        \end{itemize}
        \item Topology Evolution:
        \begin{itemize}
            \item Topology Refinement (Prompt~\ref{prompt:topology_evoluation_prompt}): Guides local graph topology evolution based on retrieved subgraphs.
        \end{itemize}
\end{itemize}

\begin{promptbox}[label={prompt:memory_unit_construction_prompt}]{Memory Construction}
\textbf{You are a memory construction agent of an agent memory system.}
You must perform two tasks:
\medskip

\textbf{Task A: Memory Unit Generation.} Based on your relevant \textbf{past memories}, create memory units for the \textbf{Latest Information}.

\textbf{Task B: Dialogue Summary Generation.}
Based on the latest information, update the  \textbf{current conversation summary}.

\medskip
The past memories include a summary of the past conversation and related conversation content. The latest information is the new input information.
When generating new memory units, you need to assess their relationship to past memories. For example, the new information might answer a question raised before, or it might serve as a supplement or extension to a past memory, and so on. You need to establish this relationship within the memory units you’ve extracted.
Use the principle of the Fewest Memory Nodes: represent the \textbf{Latest Information}  clearly and completely with the fewest nodes.

\medskip
\textbf{Task A: Memory Unit Generation}
\medskip

\textit{Step 1. Extract Information.}  
Referring to the supplementary information provided by past memories, identify the key points in the latest information and represent them using the most appropriate of the following four levels of granularity:

- entity: a referable object such as a person, place, object, or concept.

- event: a single action, interaction or occurrence, describing something that happened at a specific time

- episode: a multi-step experience with context and outcome, lasting for a period of time

- fact: a stable and general statement about somebody and something, or a universal rule and knowledge, derived or deduced from latest information

\medskip
INSTRUCTIONS:

- The past memories have already been saved and does not need to be saved again.

- Avoid mixing multiple semantics in one unit.

- It is allowed to generate multiple units if needed

- Ensure that each unit is semantically coherent and complete

\medskip
\textit{Step 2. Construct Memory Units.}  
Follow the structure below to construct memory units:

\textbf{entity:}
name; aliases (a list of alternative names, NOT pronouns); category (person | object | place | concept | etc); role (describe the role or significance of this entity).

\textbf{event:}
name (general event name); participants (a list of relevant entity names); description (a description of the event); time (exact happened time of described event: none | xxxx-xx-xx).

\textbf{episode:}
name (general episode name); description  (a description of the episode, contains the why it happened and the outcome); keywords (a list of keywords for the episode, not more than 3); time\_range (exact time range of described episode: none | xxxx-xx-xx to xxxx-xx-xx).

\textbf{fact:}
statement (a clear and universal factual statement); keywords (a list of keywords for the fact, not more than 3).

\medskip
INSTRUCTIONS:

- When declaring each field in a memory unit, you must specify the fully qualified name of the element.

- Each of these memory units should be able to be semantically understood independently.
\medskip

\textbf{Task B: Dialogue Summary Generation}

Based on the latest information, update the \textbf{current conversation summary}. This summary serves as your long-term working memory as a memory construction agent, helping you understand the context and progress of the information currently being processed. The \textbf{current conversation summary}  should include content such as how the conversation topic has evolved and any ongoing, unfinished tasks, to provide a high-level overview of the conversation’s progress.

\medskip
Keep the content concise and general; avoid using any demonstrative pronouns, and do not exceed 200 words.

\medskip
GOOD example slices:

- ``A is asking B about winter travel plans.''

- ``They are discussing John's son's academic situation.''
\end{promptbox}

\begin{promptbox}[label={prompt:memory_construction_operation_prompt}]{Node Operation Decision}
\textbf{You are maintaining a memory graph.} Your task is to decide how to handle a new memory unit.

\medskip
\textbf{Task Overview} Given a new memory unit and candidate nodes (retrieved by similarity and/or recency), you need to:

\textbf{1. Decide the operation type.}

- add: Create a new node for this memory unit

- modify: Merge this memory unit into an existing candidate node

- skip: Skip storing because an existing node already completely covers this content

\medskip
\textbf{2. Output the final node information.} (for both ``add'' and ``modify''):

- You may adjust/refine the node's metadata based on context

- For ``modify'': output the merged node information
   
- For ``add'': output the new node information (can be same as input or refined)

\medskip
\textbf{3. Identify related candidates.}

- Select candidates that have semantic and logical relationships with the final node

- These will be used for edge creation in the next step

- You do NOT need to specify edge types here, just identify which candidates are related

\medskip
\textbf{Decision Criteria}

Choose \textbf{modify} only if all of the following conditions are met:

- The new information refers to the SAME underlying entity/event/fact

- It only adds detail, clarification, or minor correction

- It does NOT introduce: a new state, a temporal change, or a contradiction

- Merging will NOT cause loss of previously stored information and the memory unit information (their content are duplicated)

\medskip
Choose \textbf{skip} if:

- An existing node already completely covers the content

- Storing this would create pure duplication

\medskip
Otherwise, choose \textbf{add}.

\medskip
\textbf{Important Principle}

When uncertain, prefer ``add'' over ``modify'', since an incorrect merge may cause irreversible information loss.

\medskip
\textbf{Candidate Sources}

Candidates come from two sources (marked in the input):

- \textbf{similarity}: Retrieved by semantic similarity to the new memory unit

- \textbf{time}: Recently modified/created nodes (temporal proximity)

- \textbf{similarity+time}: Retrieved by both methods

Consider both similarity scores and temporal context when making decisions.

\end{promptbox}

\begin{promptbox}[label={prompt:memory_relation_decision_prompt}]{Edge Connection Decision}
\textbf{You are maintaining a memory graph.} Your task is to decide how to establish edges between a target node and related nodes.

\medskip
\textbf{Task Overview}: Given a target node and its related candidate nodes (some may already have edges to the target), decide:

- For each candidate: what edge operation to perform

- Edge operations: ``add'' (new edge), ``modify'' (change existing edge), ``delete'' (remove edge), ``no\_change'' (keep as is)

\medskip
\textbf{Edge Predicates (Organized by Type)}

\textbf{1. Logical}

- \textbf{contradicts(A,B)}: A and B are logically mutually exclusive. When A is detected, the activation of B should be suppressed.

- \textbf{implies(A,B)}: If A holds, then B must hold.

- \textbf{constrains(A,B)}: The existence of A restricts the boundaries or execution parameters of B.

- \textbf{verifies(A,B)}: A serves as evidence for B, increasing confidence in B.

\medskip
\textbf{2. Causal}

- \textbf{precedes(A,B)}: Pure temporal sequence;  A occurs before B, with no direct causality.

- \textbf{enables(A,B)}: A is a necessary condition for B to occur (necessary Condition). Without A, B cannot be initiated.

- \textbf{triggers(A,B)}: The occurrence of A directly causes B (Sufficient Condition).

- \textbf{results\_in(A,B)}: B is the state change resulting from the execution of A (State Transition).

\medskip
\textbf{3. Taxonomic}

- \textbf{comprises(A,B)}: A whole-part relationship. A is composed of multiple parts, including B.

- \textbf{instantiates(A,B)}: A relationship between abstract class A and concrete instance B.

- \textbf{summarizes(A,B)}: A is a high-level semantic compression of B (typically episodes or events).

\medskip
\textbf{4. Associative}

- \textbf{correlates(A,B)}: Semantically highly similar or statistically strongly correlated.

- \textbf{supersedes(A,B)}: A is the latest version or correction of B; B should be considered obsolete or historical.

- \textbf{analogous\_to(A,B)}: An analogical relationship. A and B are similar in structure or function, although they belong to different domains.

- \textbf{contextualizes(A,B)}: A provides necessary background information or contextual clarification for B.

\medskip
\textbf{Decision Guidelines}

For add: Create a new edge only when the relationship is clear. Specify the predicate, weight, and description.

For modify: Update an existing edge when the predicate is more appropriate or the weight should be adjusted. Output the complete updated edge.

For delete: Remove an edge only when the relationship is clearly invalid. Prefer \textbf{no\_change} when uncertain.

For no\_change: Keep the existing edge unchanged when it remains valid.

\medskip
\textbf{Direction Rules}

- For single-direction predicates: source -> target indicates the relationship direction

- For bidirectional predicates (contradicts, correlates, analogous\_to): the system will automatically create reverse edges

\medskip
\textbf{Weight Guidelines}: Weight represents the certainty/strength of the relationship [0, 1]:

- 0.9-1.0: Very confident, explicit relationship

- 0.7-0.8: Confident, clear relationship

- 0.5-0.6: Moderate confidence, implicit relationship

- 0.3-0.4: Low confidence, weak relationship

- 0.0-0.2: Very uncertain
\end{promptbox}

\begin{promptbox}[label={prompt:seed_planner_prompt}]{Seed Retrieval Planner}
\textbf{Task}
Analysis the question, thinking about the key entities, contents, potential logical chains, potential related factors, etc., and generate a strategy to locate the anchor points on the memory graph.

\medskip
Choose one mode from following three modes:

- query\_match: generate a new query based on the question, and only uses embedding semantic similarity to search the memory graph. It is good at recalling memories expressed with different wording, implicit context, paraphrased situations, and broad semantic targets.

- keywords\_match: generate a list of keywords based on the question, and only uses lexical keyword matching to search the memory graph. It is good at preserving precise surface anchors such as names, places, organizations, project, object, event, and quoted phrases.

- hybrid\_match: generate a new query and a list of keywords based on the question, and take both scores into consideration to search the memory graph. It is good when initial recall should preserve exact anchors while also matching the semantic condition.

\medskip
\textbf{Return strict JSON keys:}

- allowed\_node\_types: the types of the nodes to retrieve.

-time\_range: the temporal constraint for retrieval.

\medskip
\textbf{Allowed Node Types Candidates:}

- entity: A referable object such as a person, place, object, or concept.

- event: a single action or something happened at a specific time

- episode: a multi-step experience with context and outcome, lasting for a period of time

- fact: stable, complete, generalizable and explicit statement derived or deduced from NEW CONTENT
\end{promptbox}

\begin{promptbox}[label={prompt:retrieval_sufficiency_prompt}]{Retrieval Sufficiency Controller}
You are a retrieval controller in a graph memory system. Decide whether the displayed candidate nodes already contain enough information to answer the question.

\medskip
First decide global sufficiency using all displayed candidate nodes together. Return ``is\_enough'': ``true'' when at least one candidate node, or the combination of multiple candidate nodes, contains the answer and you can directly answer the question from the shown evidence. It is NOT required that every candidate node can answer the question.
\medskip

When ``is\_enough'' is ``true'', ``nodes\_to\_expand'' must be [] and ``nodes\_to\_skip'' must be []. Do not expand other nodes just because they do not contain the answer.
\medskip

Return ``is\_enough'': ``false'' only when all displayed candidate nodes together still do not contain the answer. In that case, choose which candidate nodes should be used as starting points for the next graph expansion round.

In Graph Memory System, content-related nodes are connected by various semantic and logical relationships.

- nodes\_to\_expand: Select a node only if it is relevant to the topic of the question, contains partial evidence, provides a reasoning bridge to the answer, is located in a graph neighborhood that may contain the answer, or holds the potential to lead to evidence.
\medskip
Do not expand a node just to confirm an answer that is already explicitly stated in that node. Only select the nodes with the highest expansion potential, no more than 5 nodes.
\end{promptbox}

\begin{promptbox}[label={prompt:expansion_prompt}]{Graph Expansion Policy}
Given the current retrieval state, generate graph expansion strategy for each selected seed node. The available edge predicates are identical to those defined in the \textbf{Edge Connection Decision} prompt.

\medskip
\textbf{Strategy of Each action:}

\textbf{Necessary Parameters:}

- seed\_id: the id of the seed node.

- expansion\_mode: choose one of [``subgraph\_beam'', ``path\_search''].

\ \ \ \ - subgraph\_beam: builds a compact relevant subgraph by globally selecting high-quality candidates. Use this when the question needs broad evidence aggregation, related facts, constraints, or contextual support.
    
\ \ \ \ - path\_search: searches coherent reasoning paths. Discarded paths/candidates must not be reconsidered later. Use this when the question needs chain-like evidence, causal/logical progression, or one consistent explanation route.
    
- layer\_predicate\_include: the layer-predicate-direction constraints to include. Each item is a dictionary with the format of \{``layer'': str, ``predicate'': str, ``direction'': ``forward|backward|both|none''\}.

\ \ \ \ - layer must be one of [``logical'', ``causal'', ``taxonomic'', ``associative''].

\ \ \ \ - predicate must belong to the selected layer. If you only want to restrict by layer, use an empty predicate.

- score\_threshold: minimum candidate score to keep. Candidates below this threshold should be filtered before adding them to a subgraph or path.

\medskip
\textbf{Mode-specific Parameters:}

- For expansion\_mode = ``subgraph\_beam'':

\ \ \ \ - subgraph\_max\_nodes: maximum number of unique non-seed nodes to collect for this seed under this action. Use a small number for focused retrieval and a larger number only when the question needs broad supporting evidence.

- For expansion\_mode = ``path\_search'':

\ \ \ \ - path\_max\_depth: maximum path length measured by number of edges from the seed. This is the hard upper bound for coherent path exploration.

\ \ \ \ - path\_top\_k: maximum number of active paths to keep at each hop. Use 1 for a single greedy path, and a larger value only when multiple plausible reasoning chains should be explored.
    
\medskip
\textbf{Optional Parameters:}

- temporal\_decay: optional time constraint operator. Use this only when the question requires explicit temporal constraints.

\ \ \ \ - strict JSON format:
\small
\begin{verbatim}
  {
    ``time'': ``YYYY | YYYY-MM | YYYY-MM-DD | YYYY-MM-DD HH:MM | [start, end]'',
    ``delta_t'': ``supports value+unit, e.g. {``value'': 7, ``unit'': ``days''}'',
    ``mode'': ``hard | soft'',
    ``missing_time_mode'': ``keep | drop''
  }
\end{verbatim}
\normalsize

\ \ \ \ - semantics:

\ \ \ \ \ \ \ \ - time is determined by the question.

\ \ \ \ \ \ \ \ - delta\_t is the tolerance window (time $\pm$ delta\_t): in-window => no penalty.

\ \ \ \ \ \ \ \ - hard: out-of-window nodes should be treated as filtered.

\ \ \ \ \ \ \ \ - soft: out-of-window nodes are not filtered, but decayed by temporal penalty.

\ \ \ \ \ \ \ \ - missing\_time\_mode = keep: keep node with a fixed penalty.

\ \ \ \ \ \ \ \ - missing\_time\_mode = drop: drop node directly.

- conflict\_inhibit: the flag of the conflict inhibit.

- reasoning: describe what content this strategy is intended to find.
\end{promptbox}

\begin{promptbox}[label={prompt:topology_evoluation_prompt}]{Topology Update}
\textbf{You are a topology auditor for a graph memory system.} You are tasked with analyzing a retrieved memory subgraph and identifying how its topology should be refined.

\medskip
\textbf{Task Objectives}

This is a local topology reconstruction task. You will be provided with a set of recalled nodes and the existing edges between them. Your task is to determine whether new edges should be established between these nodes or whether the weights of existing edges should be adjusted. The purpose of these operations is to make the relationships between nodes more logically rigorous and complete, and to facilitate future retrieval.

\medskip
The provided questions and answers serve as references to assist you in reconstructing the local topology. You can analyze the contribution of these nodes to the questions to better understand the relationships between them. Your ultimate goal is not to enhance the connections within this specific problem, but to improve future graph access: key evidence should be accessible through a reliable logical structure, misleading connections should have their weights reduced, and connections between nodes unrelated to the problem should be established only if their relationships are clearly defined.

\medskip
\textbf{Context}

\textbf{Allowed Edge Schema: (Type-Predicate)}

\begin{verbatim}
{edge_type_schema}
\end{verbatim}

\medskip
\textbf{Confidence Guidelines}

For every operation, output confidence, not manual weight delta. Confidence represents how certain you are that this operation is useful and correct for future graph retrieval [0, 1]:

\begin{verbatim}
{confidence_rule}
\end{verbatim}

For add\_edge, confidence will be used directly as the new edge weight. For strengthen\_edge/weaken\_edge, the system will convert confidence to the weight change; do not output delta.

\medskip
\textbf{Required Analysis Workflow}

Before deciding edits, explicitly analyze the recalled nodes in the JSON output:

1. Determine the core topic represented by the recalled nodes. The input question should only be treated as a reference signal for identifying the topic. If the answer to the question is incorrect, focus instead on the underlying topic that the recalled nodes collectively imply.

2. Identify which nodes constitute the key evidence required to answer the question correctly, or are directly relevant to the identified topic. Input them at ``gold\_supporting\_node\_ids''.

3. Identify which nodes provide supporting context, including but not limited to: background information, constraints, temporal information, entity disambiguation, or other information necessary for understanding the topic. Input them at ``supporting\_associated\_node\_ids''.

4. Identify which nodes are irrelevant to the answer or topic, or may introduce misleading directions during retrieval and reasoning. Input them at ``misleading\_node\_ids'' or ``unrelated\_node\_ids''.

5. Based on the above analysis, propose topology refinement actions. In particular:

\ \ - strengthen predicate relations among generally relevant nodes,

\ \ - weaken the influence of topic-irrelevant or misleading nodes,

\ \ - add missing connections between nodes that should be related through valid predicates but are currently disconnected,

\ \ - improve the logical coherence of the local subgraph to better support future retrieval and reasoning.

\medskip
\textbf{Constrains}
1. Only create new edges between recalled nodes, and only adjust the weights of existing edges between these nodes.

2. Preserve the existing graph structure. Do not delete edges or modify node content.

3. Use only predicates from the provided allowed edge predicate set.

4. When adjusting an existing relationship, use strengthen\_edge or weaken\_edge and reference a valid edge\_id from the existing graph.

5. Add a new edge only when the relationship is strongly supported by the available evidence from the recalled nodes. If the evidence is ambiguous, do not add the edge.

6. Adjust edge weights only when there is clear evidence that the relationship strength should change and that doing so is likely to improve future memory retrieval.
\end{promptbox}


\end{document}